%% file: eswc22-clp-frame.tex
\documentclass[runningheads]{llncs}

\usepackage{amsmath}
\usepackage{amssymb}

\usepackage[numbers,sort]{natbib} 

\usepackage{booktabs}
\usepackage{colortbl}

\usepackage{float}
\usepackage{subcaption}
\usepackage{color}
\usepackage{dsfont}
\usepackage{algorithm}
\usepackage{algorithmic}

\usepackage{times}
\usepackage{microtype}

\usepackage{tikz}
\usetikzlibrary{backgrounds}
\usetikzlibrary{topaths,calc}
\usetikzlibrary{shadows}
\usetikzlibrary{decorations.pathreplacing}
\usetikzlibrary{shapes,positioning}
\usetikzlibrary{arrows.meta,arrows}
\usetikzlibrary{shapes.misc}
\tikzset{resource/.style={draw,rectangle,fill=lightblue,rounded corners,inner sep=5pt} }
\tikzset{literal/.style={draw,rectangle,fill=lightgreen,inner sep=5pt}}
\tikzset{>=triangle 45}
\tikzset{every picture/.style=thick}
\tikzstyle{every node}=[font=\small]

\usepackage[bookmarksnumbered,unicode]{hyperref}
\hypersetup{hidelinks}
\usepackage{color}

\begin{document}

\input{eswc22-clp-pre}
\input{eswc22-clp-part1}
\input{eswc22-clp-part2}
\input{eswc22-clp-part3}
\input{eswc22-clp-part4}
\input{eswc22-clp-part5}
\input{eswc22-clp-part6}
\input{eswc22-clp-sum}

\bibliographystyle{splncs04nat}
\bibliography{eswc22-clp-frame}

\end{document}

%% file: eswc22-clp-pre.tex
\author{N'Dah Jean Kouagou\orcidID{0000-0002-4217-897X} \and
Stefan Heindorf\orcidID{0000-0002-4525-6865} \and
Caglar Demir\orcidID{0000-0001-8970-3850} \and
Axel-Cyrille Ngonga Ngomo\orcidID{0000-0001-7112-3516}}
\authorrunning{N. J. Kouagou et al.}
\institute{Paderborn University, Paderborn, Germany\\
\email{nkouagou@mail.uni-paderborn.de, heindorf@uni-paderborn.de, caglar.demir@upb.de, axel.ngonga@upb.de}\\}
\title{Learning Concept Lengths Accelerates \texorpdfstring{\\}{} Concept  Learning in \texorpdfstring{$\mathcal{ALC}$}{ALC}}
\maketitle
\begin{abstract}
Concept learning approaches based on refinement operators explore partially ordered solution spaces to compute concepts, which are used as binary classification models for individuals. However, the number of concepts explored by these approaches can grow to the millions for complex learning problems. This often leads to impractical runtimes. We propose to alleviate this problem by predicting the length of target concepts before the exploration of the solution space. By these means, we can prune the search space during concept learning. To achieve this goal, we compare four neural architectures and evaluate them on four benchmarks. Our evaluation results suggest that recurrent neural network architectures perform best at concept length prediction with a macro F-measure ranging from 38\% to 92\%. We then extend the CELOE algorithm, which learns $\mathcal{ALC}$ concepts, with our concept length predictor. Our extension yields the algorithm CLIP. In our experiments, CLIP is at least 7.5$\times$ faster than other state-of-the-art concept learning algorithms for $\mathcal{ALC}$---including CELOE---and achieves significant improvements in the F-measure of the concepts learned on 3 out of 4 datasets. For reproducibility, we provide our implementation in the public GitHub repository at \url{https://github.com/dice-group/LearnALCLengths}.

\keywords{Concept learning \and Concept length \and Structured machine learning \and Description logic \and Learning from examples \and Prediction of concept lengths}
\end{abstract}

%% file: eswc22-clp-part1.tex
\section{Introduction}
Knowledge bases have recently become indispensable in a number of applications driven by machine learning~\cite{deshpande2013building}.
For instance, the Gene Ontology (GO)~\cite{ashburner2000gene,gene2004gene}, DrugBank~\cite{wishart2018drugbank}, and the Global Network of Biomedical Relationships (GNBR)~\cite{percha2018global} are actively being used to find treatments for certain diseases~\cite{ioannidis2020drkg,maclean2021knowledge}.
We consider the supervised machine learning task of concept learning\footnote{Also called class expression learning (CEL) \cite{fanizzi2008dl}. See Section \ref{sec:background} for a formal definition.}
\cite{lehmann2010concept}
on knowledge bases in the description logic (DL) $\mathcal{ALC}$ (attributive language with complements)~\cite{schmidt1991attributive}. We focus on approaches based on refinement operators~\cite{lehmann2010concept,badea2000refinement,fanizzi2008dl,rizzo2018framework,rizzo2020class}. 

Recent works on concept learning over DLs~\cite{bin2016towards,heindorf2021evolearner,rizzo2020class} indicate that approaches based on refinement operators often fail to achieve practical runtimes on large real-world knowledge bases, which often contain millions of individuals and concepts with billions of assertions. 
As noted by \citet{rizzo2020class}, this is partially due to the size of the search space that needs to be explored to detect relevant concepts.
In this paper, we \textit{accelerate concept learning by predicting the length of the target concept} in advance. By these means, we can prune the search space traversed by a refinement operator and therewith reduce the overall runtime of the concept learning process. To quantify our runtime improvement, we compare our new algorithm---dubbed CLIP---against the state-of-the-art approaches CELOE~\cite{lehmann2011class}, OCEL~\cite{lehmann2010concept}, and ELTL~\cite{buhmann2016dl}. 
The price for our runtime improvement is paid in the prior training of the concept length predictor. Therefore, we also show that the prediction of concept lengths can be carried out using rather simple neural architectures.

To the best of our knowledge, no similar work has been carried out before. Hence, we hope that our findings will serve as a foundation for more investigations in this direction. In a nutshell, our contributions are as follows:
\begin{enumerate}
    \item We design different neural network architectures for learning concept lengths.
    \item We implement a length-based refinement operator to generate training data.
    \item We integrate our concept length predictors into the CELOE algorithm, resulting in a new algorithm that we call CLIP. We show that CLIP achieves state-of-the-art performance in terms of F-measure while outperforming the state of the art in terms of runtime.
\end{enumerate}

The remainder of the paper is organized as follows: In Section~\ref{sec:background}, we give a brief overview of the required background in DL, concept learning, knowledge graph embeddings, and refinement operators for DLs. We also present the notation and terminology used in the rest of the paper.
Section~\ref{sec:related-work} presents related work on concept learning using refinement operators. In Sections~\ref{sec:method} and~\ref{sec:clip}, we describe our new approach for concept learning in $\mathcal{ALC}$. Our results on different knowledge bases are presented in Section~\ref{sec:evaluation}. Section~\ref{sec:conclusion} draws conclusions from our findings and introduces new directions for future work.

%% file: eswc22-clp-part2.tex
\section{Background}
\label{sec:background}
In this section, we present the background on description logics, concept learning, refinement operators, and knowledge graph embeddings. We also introduce the notation and terminology used throughout the paper. 

\subsubsection{Description Logics.}
\label{syntactic-length}
Description logics~\cite{baader2003handbook} are a family of languages for knowledge representation.
While there are more powerful variants of description logics~\cite{kroetzsch2012description,baader2003handbook}, we focus on the description logic $\mathcal{ALC}$  ($\mathcal{A}$ttributive $\mathcal{L}$anguage with $\mathcal{C}$omplement) because it is the simplest closed description logic with respect to propositional logics.
Its basic components are \textit{concept} names (e.g., $\mathit{Teacher}$, $\mathit{Human}$), \textit{role} names (e.g., $\mathit{hasChild}$, $\mathit{bornIn}$) and \textit{individuals} (e.g., $\mathit{Mike}$, $\mathit{Jack}$). Table~\ref{tab:alc-syntax-semantics} introduces the syntax and semantics of $\mathcal{ALC}$ (see \cite{lehmann2010concept} for more details).
\input{table-alc-syntax-semantics}
In $\mathcal{ALC}$, concept lengths are defined recursively \cite{lehmann2010concept}
\begin{enumerate}
    \item $\mathit{length}(A) = \mathit{length}(\top) = \mathit{length}(\bot) = 1$, for all atomic concepts $A$
    \item $\mathit{length}(\neg C) = 1 + \mathit{length}(C)$, for all concepts $C$
    \item $\mathit{length}(\exists~ r.C) = \mathit{length}(\forall~ r.C) = 2 + \mathit{length}(C)$, for all concepts $C$
    \item $\mathit{length}(C\hspace{0.05cm}\sqcup\hspace{0.05cm}D) = \mathit{length}(C\hspace{0.05cm}\sqcap\hspace{0.05cm} D) = 1 + \mathit{length}(C) + \mathit{length}(D)$, for all concepts $C$ and $D$.
\end{enumerate}
The pair $\mathcal{K}=(\mathit{TBox}, \mathit{ABox})$ denotes an $\mathcal{ALC}$ knowledge base. The $\mathit{TBox}$ contains statements of the form $C\sqsubseteq D$ or $C\equiv D$, where $C$ and $D$ are concepts. The $\mathit{ABox}$ consists of statements of the form $C(a)$ and $R(a,b)$, where $C$ is a concept, $R$ is a role, and $a, b$ are individuals. $N_C$ and $N_R$ are the sets of concept names and role names in~$\mathcal{K}$, respectively. $\mathcal{K}_I$ stands for the set of all individuals in $\mathcal{K}$. $|.|$ denotes the cardinality function, that is, a function that takes a set as input and returns the number of elements in the set. Given a concept $C$, we denote the set of all instances of $C$ by~$C_\mathcal{P}$. $C_\mathcal{N}$ stands for the set of all individuals that are not instances of~$C$.

\subsubsection{Concept Learning.}
We recall the definition introduced by \cite{lehmann2010concept}.
\begin{definition}
\label{definition-concept-learning-exact}
Let $\mathcal{K}$, $T$, $P$, and $N$ be a knowledge base, a target concept, and sets of positive and negative examples from $\mathcal{K}_I$, respectively.
The learning problem is to find a concept $C$ such that $T$ does not occur in $C$, and for $\mathcal{K}' = \mathcal{K} \cup \{T\equiv C\}$, we have that $\mathcal{K}' \models P$ and $\mathcal{K}' \not \models N$.
\end{definition}
Since such a concept $C$ does not always exist, we target an approximate definition in this work. 
\begin{definition}
\label{definition-concept-learning-approximate}
Given a knowledge base $\mathcal{K}$, a set of positive examples $P$, and a set of negative examples $N$, the learning problem is to find a concept $C$ which maximizes the F-measure $F$, where $F$ is defined by
$
    F  =  2\times \frac{Precision \times Recall}{Precision + Recall}, \text{ with }
    Precision = \frac{|C_\mathcal{P}\cap P|}{|C_\mathcal{P}\cap P|+|C_\mathcal{P}\cap N|} \text{ and } 
    Recall = \frac{|C_\mathcal{P}\cap P|}{|C_\mathcal{P}\cap P|+|C_\mathcal{N} \cap P|}.
$
\end{definition}
Note that in the above definition, $C_\mathcal{N}$ and $C_\mathcal{P}$ depend on both the concept $C$ and the learning problem. Following \cite{lehmann2010concept}, we use the closed-world assumption (CWA) to compute $C_\mathcal{P}$ and $C_\mathcal{N}$: every individual that cannot be inferred to be an element of $C_\mathcal{P}$ is considered to be in $C_\mathcal{N}$. In this work, we are interested in finding such a concept $C$ by using refinement operators (see the following subsection). The found concept~$C$ might not be unique for a learning problem. 

\subsubsection{Refinement Operators.}
\begin{definition}
\label{definition-refinement-operators}
A quasi-ordering $\preceq$ is a reflexive and transitive binary relation. Let \mbox{$(\mathcal{S}, \preceq)$} be a quasi-ordered space. A downward (upward) refinement operator on $\mathcal{S}$ is a mapping $\rho: \mathcal{S}\rightarrow 2^\mathcal{S}$ such that for all $C\in \mathcal{S}$, $C' \in \rho(C)$ implies  $C'\preceq C$ $(C\preceq C')$.
\end{definition}
\begin{example}
Let $\mathcal{K}=(\mathit{TBox}, \mathit{ABox})$ be a knowledge base, with 
\begin{align*} 
\mathit{TBox} &=\{\mathit{Female} \sqsubseteq \mathit{Human}, \mathit{Mother}\sqsubseteq \mathit{Female}, \mathit{Human} \sqsubseteq \neg \mathit{Car}\};\\
\mathit{ABox} &=\{\mathit{Female(Anna)}, \mathit{Mother(Kate)}, \mathit{Car(Venza)}, \mathit{hasChild(Jack, Paul)}\}.
\end{align*} 
Assume the sets of concept names $N_C$ and role names $N_R$ in $\mathcal{K}$ are given by:
\begin{align*} 
N_C&=\{\mathit{Car}, \mathit{Female}, \mathit{Human}, \mathit{Mother}, \mathit{Parent}\}; \\ N_R&=\{\mathit{hasChild}, \mathit{manufacturedBy}, \mathit{marriedTo}\}.
\end{align*}
Let $\mathbf{C}$ be the set of all $\mathcal{ALC}$ concept expressions \cite{rudolph2011foundations} that can be constructed from $N_C$ and $N_R$ (note that $\mathbf{C}$ is infinite and every concept name is a concept expression).
Consider the mapping $\rho: \mathbf{C}\rightarrow 2^\mathbf{C}$ defined by: $\rho(C)=\{C'\in \mathbf{C}|C'\sqsubseteq C, C' \neq C\}$ for all $C\in\mathbf{C}$. $\rho$ is clearly a downward refinement operator and we have for example,
\begin{itemize}
  \item $\{\mathit{Female}, \mathit{Mother}, \mathit{Female} \sqcup \mathit{Mother}\}\subseteq \rho(Human)$;
  \item $\{\exists~\mathit{marriedTo.Mother}, \forall~\mathit{marriedTo.Female}\}\subseteq \rho(\exists~\mathit{marriedTo}.(\neg \mathit{Car}))$.
\end{itemize}
\end{example}

Refinement operators can have a number of important properties which we do not discuss in this paper (for further details, we refer the reader to \cite{lehmann2010concept}). In the context of concept learning, these properties can be exploited to optimize the traversal of the concept space in search of a target concept.

\subsubsection{Knowledge Graph Embeddings.}
The $\mathit{ABox}$ of a knowledge base in $\mathcal{ALC}$ can be regarded as a knowledge graph~\cite{hogan2021knowledge} (see also Section \ref{sec:method}). A knowledge graph embedding function typically maps a knowledge graph to a continuous vector space to facilitate downstream tasks such as link prediction and knowledge graph completion~\cite{wang2017knowledge,dai2020survey}. We exploit knowledge graph embeddings to improve concept learning in DLs. Knowledge graph embedding approaches can be subdivided into two categories: the first category of approaches uses only facts in the knowledge graph~\cite{nickel2012factorizing,weston2013connecting,bordes2014semantic}, and the second category of approaches takes into account additional information about entities and relations, such as textual descriptions~\cite{xie2016representation,wang2016text}. Both approaches typically initialize each entity and relation with a random vector, matrix, or tensor. Then, a scoring function is defined  to learn embeddings so that facts observed in the knowledge graph receive high scores, while unobserved facts receive low scores. It can also happen that unobserved facts receive a high score, for instance, if a fact is supposed to hold but it is not observed in the knowledge graph, or if it is a logical implication of the learned patterns. For more details, we refer the reader to the surveys~\cite{wang2017knowledge,dai2020survey}. In this work, we use the Convolutional Complex Embedding Model (ConEx) \cite{demir2020convolutional}, which has been shown to produce state-of-the-art results with fewer trainable parameters.

%% file: table-alc-syntax-semantics.tex
\begin{table}[tb]
	\centering
	\caption{$\mathcal{ALC}$ syntax and semantics. $\mathcal{I}$ stands for an interpretation, $\Delta^\mathcal{I}$ for its domain.}
	\label{tab:alc-syntax-semantics}
	\setlength{\tabcolsep}{10pt}
   \begin{tabular}{@{}lll@{}}
        \toprule
		\textbf{Construct} & \textbf{Syntax} & \textbf{Semantics}\\
		\midrule
        Atomic concept & $A$ & $A^{\mathcal{I}}\subseteq{\Delta^\mathcal{I}}$\\
        Atomic role & $r$ & 
		$r^\mathcal{I}\subseteq{\Delta^\mathcal{I}\times \Delta^\mathcal{I}}$\\
		Top concept & $\top$ & $\Delta^\mathcal{I}$\\
		Bottom concept & $\bot$& $\emptyset$\\
		Conjunction & $C\sqcap D$& $C^\mathcal{I}\cap D^\mathcal{I}$\\
		Disjunction & $C\sqcup D$& $C^\mathcal{I}\cup D^\mathcal{I}$\\
		Negation & $\neg C$ & $\Delta^\mathcal{I}\setminus C^\mathcal{I}$\\
		Existential restriction & $\exists~ r.C$ & $\{a^\mathcal{I}/\exists~ b^\mathcal{I} \in C^\mathcal{I}, (a^\mathcal{I},b^\mathcal{I})\in r^\mathcal{I}\}$\\
		Universal restriction & $ \forall~ r.C$ & $\{a^\mathcal{I}/\forall~ b^\mathcal{I}, (a^\mathcal{I},b^\mathcal{I})\in r^\mathcal{I}\Rightarrow b^\mathcal{I}\in C^\mathcal{I}\}$\\
		\bottomrule
	\end{tabular}
\end{table}

%% file: eswc22-clp-part3.tex
\section{Related Work}
\label{sec:related-work}
Lehmann and Hitzler \cite{lehmann2010concept} investigated concept learning using refinement operators by studying combinations of possible refinement operator properties and designed their own refinement operator in $\mathcal{ALC}$.
Their approach proved to be competitive in accuracy with (and in some cases, superior to) the state-of-the-art, namely inductive logic programs.
Badea and Nienhuys-Cheng \cite{badea2000refinement} worked on a similar topic in the DL $\mathcal{ALER}$.
The evaluation of their approach on real ontologies from different domains showed promising results, but it had the disadvantage of depending on the instance data.

DL-Learner~\cite{lehmann2009dl} is the most mature framework for concept learning. CELOE~\cite{lehmann2011class}, OCEL~\cite{lehmann2010learning}, and ELTL~\cite{buhmann2016dl} are algorithms implemented in DL-Learner. CELOE is an extension of OCEL that uses the same refinement operator but with a different heuristic function. It is considered the best algorithm in DL-Learner to date. The algorithm uses a soft syntactic bias in its heuristic function that balances between predictive performance and short, readable concepts. In this work, we opt for a hard syntactic bias that constrains our algorithm to generate expressions shorter than a given threshold. ELTL is designed for the simple description logic $\mathcal{EL}$.
Despite its usefulness, DL-Learner suffers from performance issues in certain scenarios~\cite[cf.][]{heindorf2021evolearner,sarker2019efficient}.

\textsc{DL-Foil} \cite{fanizzi2008dl,rizzo2018framework} is another concept learning algorithm that uses refinement operators and progressively constructs the solution as a disjunction of partial descriptions. Each partial description covers a part of the positive examples and rules out as many negative/uncertain-membership examples as possible.
\textsc{DL-Focl1--3} \cite{rizzo2020class} are variants of \textsc{DL-Foil} that employ meta-heuristics to help reduce the search space.
The first release of \textsc{DL-Focl}, also known as \textsc{DL-Focl1}, is essentially based on omission rates: to check if further iterations are required, \textsc{DL-Focl1} compares the score of the current concept definition with that of the best concept obtained at that stage. \textsc{DL-Focl2} employs a look-ahead strategy by assessing the quality of the next possible refinements of the current partial description. Finally, \textsc{DL-Focl3} attempts to solve the myopia problem in \textsc{DL-Foil} by introducing a local memory, used to avoid reconsidering sub-optimal choices previously made. EvoLearner~\cite{heindorf2021evolearner} is a concept learner based on evolutionary algorithms. In contrast, we propose to learn \emph{concept lengths} from positive and negative examples to boost the performance of \emph{concept learners}. 

%% file: eswc22-clp-part4.tex
\section{Concept Length Prediction}
\label{sec:method}

In this section, we address the following learning problem: Given a knowledge base $\mathcal{K}$, a set of positive examples $P$ and negative examples $N$, predict the length of the shortest concept $C$ that is a solution to the learning problem defined by $\mathcal{K}$, $P$, and $N$ according to Definition \ref{definition-concept-learning-approximate}. To achieve this goal, we devise a generator that creates training data for our prediction algorithm based solely on $\mathcal{K}$ and a user-given number of learning problems to use at training time. 
\subsection{Training Data for Length Prediction}
\subsubsection{Data Generation.}
\label{subsubsec:data-generation}
Given a knowledge base, the construction of training data (concepts with their positive and negative examples) is carried out as follows:
\begin{enumerate}
    \item \label{short} Generate concepts of various lengths using the length-based refinement operator described in Algorithm~\ref{alg:get} and \ref{alg:refine}. In this process, short concepts are preferred over long concepts, i.e., when two concepts have the same set of instances, the longest concept is left out.
    \item Compute the sets $C_\mathcal{N}$ and $C_\mathcal{P}$ for each generated concept $C$.
    \item Define a hyper-parameter $\mathbf{N} \in [1, |\mathcal{K}_I|]$ that represents the total number of positive and negative examples we want to use per learning problem. 
    \item Sample positive and negative examples as follows:
        \begin{itemize}
            \label{sampling}
            \item[$\bullet$] If $|C_{\mathcal{P}}|\ge \frac{\mathbf{N}}{2}$ and $|C_{\mathcal{N}}|\ge \frac{\mathbf{N}}{2}$, then we randomly sample $\frac{\mathbf{N}}{2}$ individuals from each of the two sets $C_{\mathcal{P}}$ and $C_{\mathcal{N}}$.
            \item[$\bullet$] Otherwise, we take all individuals in the minority set and sample the remaining number of individuals from the other set.
        \end{itemize}
\end{enumerate}

\subsubsection{Training Data Features.}
A knowledge graph is commonly defined as $\mathcal{G} \subseteq \mathcal{E} \times \mathcal{R} \times \mathcal{E}$, where $\mathcal{E}$ is a set of entities and $\mathcal{R}$ is a set of relations. We convert a given knowledge base $\mathcal{K}$ into a knowledge graph by converting $\mathit{ABox}$ statements of the form $R(a, b)$  into $(a, R, b)$. Statements of the form $C(a)$ are converted into $(a, \texttt{rdf:type}, C)$. 
In our experimental data, the $\mathit{TBox}$es contained only subsumptions $C \sqsubseteq D$ between atomic concepts $C$ and $D$, which were converted into triples $(C, \texttt{rdfs:subClassOf},  D)$. Hence, in our experiments, $\mathcal{E} \subseteq N_C \cup \mathcal{K}_I$ and $\mathcal{R} = N_R \cup \{\texttt{rdfs:subClassOf}\}$.

The resulting knowledge graph is then embedded into a continuous vector space to serve for the prediction of concept lengths.
On the vector representation of entities, we create an extra dimension at the end of the entries, where we insert $+1$ for positive examples and $-1$ for negative examples. Formally, we define an injective function $f_C$ for each target concept $C$
\begin{align}
    \nonumber &f_C: \mathbb{R}^\textbf{d} \longrightarrow \mathbb{R}^{\textbf{d}+1}\\
    &\mathbf{x}=(x_1, \dots, x_\textbf{d}) \longmapsto \begin{cases} (x_1, \dots, x_\textbf{d}, \phantom{-}1) & \text{if } \mathit{ent}(\mathbf{x}) \in C_\mathcal{P},\\
    (x_1, \dots, x_\textbf{d}, -1) &\text{otherwise},\end{cases}
\end{align}
where $\textbf{d}$ is the dimension of the embedding space, and $\mathit{ent}(\mathbf{x})$ is the entity whose embedding is $\mathbf{x}$. 
Thus, a data point in the training, validation, and test datasets is a tuple $(M_C, \mathit{length}(C))$, where $M_C$ is a matrix of shape $\textbf{N}\times(\textbf{d}+1)$ constructed by concatenating the embeddings of positive examples followed by those of negative examples. Formally, assume $n_1 \text{ and } n_2$ are the numbers of positive and negative examples for $C$, respectively. Further, assume that the embedding vectors of positive examples are $x^{(i)}, ~i=1\dots, n_1$ and those of negative examples are $x^{(i)},~i=n_1+1,\dots, n_1+n_2=\textbf{N}$. Then, the $i-th$ row of $M_C$ is given by
\begin{align}
    \label{features}
    M_C[i,:] = \begin{cases} (x_1^{(i)}, \dots, x_\textbf{d}^{(i)}, \phantom{-}1) & \text{if } 1\le i \le n_1,\\
    (x_1^{(i)}, \dots, x_\textbf{d}^{(i)}, -1) & \text{if } n_1+1\le i \le n_1+n_2=\mathbf{N}.
    \end{cases}
\end{align}

\input{table-dataset-concepts}

We view the prediction of concept lengths as a classification problem with classes $0, 1, \dots, L$, where $L$ is the length of the longest concept in the training dataset. As shown in Table \ref{tab:dataset-concepts}, the concept length distribution can be imbalanced. To prevent concept length predictors from overfitting on the majority classes, we used the weighted cross-entropy loss
\begin{align}\mathcal{L}_w(\bar{y}, y)=-\frac{1}{bs}\sum_{i=1}^{bs}\sum_{k=1}^{L}w_k\mathds{1}{(k,y^i)}\log(\bar{y}_k^i),\end{align} where $bs$ is the batch size, $\bar{y}$ is the batch matrix of predicted probabilities (or scores), $y$ is the batch vector of targets, $w$ is a weight vector, and $\mathds{1}$ is the indicator function. 
The weight vector is defined by: $w_k=1 /\sqrt{[k]}$, where $[k]$ is the number of concepts of length $k$ in the training dataset.
Table \ref{tab:dataset-concepts} provides details on the training, validation, and test datasets for each of the four knowledge bases. Though the maximal length for the generation of concepts was fixed to 15, many long concepts were equivalent to shorter concepts. As a result, they were removed from the training dataset and the longest remaining are of length 11 (see Table \ref{tab:dataset-concepts}).

\subsection{Concept Length Predictors}
We consider four neural network architectures: Long Short-Term Memory (LSTM) \cite{hochreiter1997long}, Gated Recurrent Unit (GRU) \cite{cho2014learning}, Multi-Layer Perceptron (MLP), and Convolutional Neural Network (CNN). Recurrent neural networks (LSTM, GRU) take as input a sequence of embeddings of the positive and negative examples (all positive examples followed by all negative examples). The CNN model takes the same input as recurrent networks and views it as an image with a single channel. In contrast, the MLP model inputs the average embeddings of a set of positive and negative examples. The implementation details and the hyper-parameter setting for each of the networks are given in Section \ref{hyper-parameter}.

%% file: table-dataset-concepts.tex
\begin{table}[tb]
	\centering
	\caption{Number of concepts per length in the training, validation, and test datasets for the four knowledge bases considered}
	\label{tab:dataset-concepts}
	\setlength{\tabcolsep}{4.3pt}
	\begin{tabular}{@{}crrrrrrrrrrrr@{}}
		\toprule
		 \textbf{Length} & 
		\multicolumn{3}{c}{\textbf{Carcinogenesis}} & \multicolumn{3}{c}{\textbf{Mutagenesis}} & \multicolumn{3}{c}{\textbf{Semantic Bible}} &
		\multicolumn{3}{c}{\textbf{Vicodi}}\\
		\cmidrule(lr){2-4} 
		\cmidrule(lr){5-7} 
		\cmidrule(lr){8-10}
		\cmidrule(l){11-13}
		& \textbf{Train} & \textbf{Val.} & \textbf{Test} & \textbf{Train} & \textbf{Val.} & \textbf{Test} & \textbf{Train} & \textbf{Val.} & \textbf{Test} & \textbf{Train} & \textbf{Val.} & \textbf{Test}\\
	\midrule 
		3 & 3,647 &  405  &  1,013  &   1,038 & 115 & 288   & 487 & 54 & 135 & 3,952 & 439 & 1,098\\
		5 &  782 &   87  &   217  &   1,156 & 129 & 321   & 546 & 61 & 152 & 2,498 & 278 & 694\\
		6 &   0 &   0  &    0  &     0 &   0 &   0   & 162 &  18 & 45 & 335 & 37 & 93\\
		7 & 1,143 &  127  &  318 & 1,310 & 146 & 364   & 104 & 12 & 29 & 3,597 & 400 & 999\\
		8 &    0 &    0  &     0  & 0 &   0 &   0   &  0 &  0 & 0 & 747 & 83 & 207\\
		9 &    0 &    0  &     0  & 0 &   0 &   0   &  73 &  8 & 21 & 0 & 0 & 0\\ 
		
		11 &   0 &    0  &     0   & 0 &      0 &   0 &   41  &  5 &  11 & 0 & 0 & 0\\
		\bottomrule
	\end{tabular}
\end{table}

%% file: eswc22-clp-part5.tex
\section{Concept Learner with Integrated Length Prediction (CLIP)}
\label{sec:clip}
The intuition behind CLIP is that \emph{if we have a reliable concept length predictor, then our concept learner only needs to test concepts of length up to the predicted length}.
Figure \ref{fig:celoe-clp} illustrates CLIP's exploration strategy.
\input{figure-clip-exploration}

Refinements that exceed the predicted length are ignored during the search. In the figure,  the concept $\texttt{Person}$ $\sqcap$ $(\forall~\texttt{attendsSome.(Workshop $\sqcup$ Conference))}$ is of length 7 and is therefore  neither tested nor added to the search tree. 
 \begin{remark}
 \label{rem2}
 For concept length prediction during concept learning, we sample $n_1$ positive examples and $n_2$ negative examples from the considered learning problem such that $n_1+n_2=\textbf{N}$, as described in Section~\ref{subsubsec:data-generation} (\ref{sampling}).
 \end{remark}
We implemented the intuition behind CLIP by extending CELOE's refinement operator.
Our refinement operator differs from CELOE's in how it refines atomic concepts (see Algorithms \ref{alg:get} and \ref{alg:refine}). For example, it considers all refinements $A' \sqsubset A$ of an atomic concept $A$ whereas CELOE's refinement operator only considers $A' \sqsubset A$ such that there is no $A''$ with $A' \sqsubset A'' \sqsubset A$. Omitting this expensive check allows more concepts to be tested in the same amount of time. In the following, we describe our method for refining atomic concepts and refer the reader to \cite{lehmann2010concept,lehmann2011class} for details on CELOE. 
In Algorithms \ref{alg:get} and \ref{alg:refine}, the hyper-parameters $\mathit{max\_length}$, $k$, and $\mathit{construct\_frac}$ control the refinement operator: $\mathit{max\_length}$ specifies how long the refinements can become (Algorithm~\ref{alg:refine}, lines 8, 11, 13); $k$ controls the number of fillers sampled without replacement for universal and existential restrictions (Algorithm~\ref{alg:get}, line 7); $\mathit{construct\_frac} \in (0,1]$ specifies the fraction of constructs to be sampled (Algorithm~\ref{alg:refine}, lines 1--3).

Given a knowledge base $\mathcal{K}$, the refinement of an atomic concept $A$ is carried out as follows: (1) obtain the subconcepts $\mathit{subs}$ of $A$ in $\mathcal{K}$; (2) compute the negations $\mathit{neg\_subs}$ of all subconcepts of $A$; (3) construct existential and universal role $\mathit{restrictions}$ where the fillers are in the set made of $\top, \bot, A$, and sample $k$ elements from each of the sets $\mathit{subs}$ and $\mathit{neg\_subs}$; (4) obtain the union $\mathit{constructs}$ of $\mathit{subs}$, $\mathit{neg\_subs}$, and $\mathit{restrictions}$, and finally (5) Algorithm~\ref{alg:refine} returns the refinements as intersections or unions of the $subs$ and $\mathit{constructs}$ computed before, with the generated refinements having length at most $\mathit{max\_length}$. The refinement operator is designed to yield numerous meaningful downward refinements from a single atomic concept.
\input{algorithm-refine-atomic-concept}

%% file: figure-clip-exploration.tex
 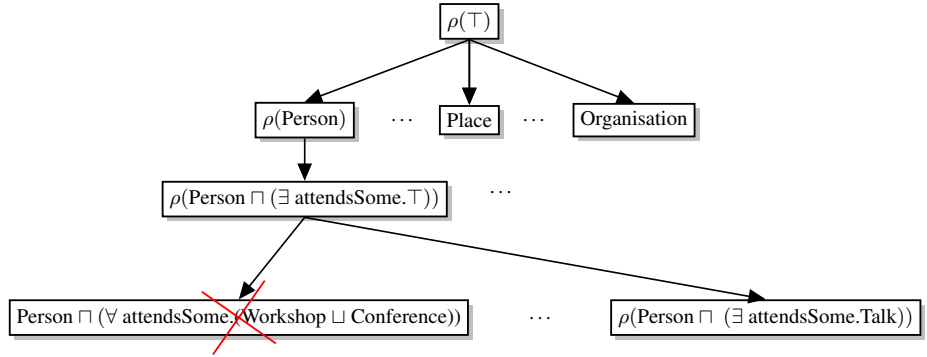
\begin{figure}[tb]
 \centering
 \resizebox{\textwidth}{!}{
 \begin{tikzpicture}[scale=1.0,baseline,thick]
	\tikzset{vertex/.style ={draw=black,shape=rectangle,fill=white,minimum size=12pt,drop shadow}}
	\tikzset{cross/.style={cross out, draw=red, minimum size=3*(#1-\pgflinewidth), inner sep=0pt, outer sep=0pt}, cross/.default={10pt}}
	\node at (2.5,4.5)[vertex] (thing) {$\rho(\top)$};      
	\node at (0,3)[vertex] (person) {$\rho(\text{Person})$}; 
	\node at (1.5,3) (subl) {\dots};
	\node at (2.5,3)[vertex] (place) {Place};
	\node at (3.5,3) (subl) {\dots};
	\node at (5,3)[vertex] (organisation) {Organisation};
	\node at (0,1.8)[vertex] (personthatworks) {$\rho(\text{Person}\sqcap(\exists~\text{attendsSome.}\top))$};
	\node at (3,1.9) (dots) {\dots};
	\node at (-1,0)[vertex] (parentOfLeafs) {$\text{Person}\sqcap(\forall~\text{attendsSome.(Workshop $\sqcup$ Conference))}$};
	\node at (3.6,0) (subl) {\dots};
	\node at (7.,0)[vertex] (personAndAtLeastLikesLiterature) {$\rho(\text{Person} \sqcap\  (\exists~ \text{attendsSome.Talk}))$};
	\tikzstyle{EdgeStyle}=[->,>=stealth,thick]
	\draw[->] [line width=0.25mm ] (thing.south) -- (person.north);
	\draw (-1,0) node[cross,rotate=10] {};
	\draw[->] [line width=0.25mm ] (thing.south) -- (place.north);
	\draw[->] [line width=0.25mm ] (thing.south) -- (place.north);
	\draw[->] [line width=0.25mm ] (thing.south) -- (organisation.north);
	\draw[->] [line width=0.25mm ] (person.south) -- (personthatworks.north);
	\draw[->] [line width=0.25mm ] (personthatworks.south) -- (parentOfLeafs.north);
	\draw[->] [line width=0.25mm ] (personthatworks.south) -- (personAndAtLeastLikesLiterature.north);
	\end{tikzpicture}}
	\caption{CLIP search tree when the predicted length is 5. After each refinement, CLIP discards all concepts whose length is larger than the set threshold.}
	\label{fig:celoe-clp}
\end{figure}

%% file: algorithm-refine-atomic-concept.tex
\begin{algorithm}[p]
	\caption{Function \textsc{RefineHelper}}
	\label{alg:get}
	\textbf{Input}: Knowledge base $\mathcal{K}$, atomic concept $A$\\
	\textbf{Hyper-parameters}: Number of subconcepts to be sampled: $k$, default 5\\
	\textbf{Output}: Subconcepts, negated subconcepts and restrictions of $A$
	\begin{algorithmic}[1]
	    \STATE subs $\leftarrow$ \textsc{Subconcepts}$_\mathcal{K}(A)$ \quad \# \textit{Subconcepts of $A$ in $\mathcal{K}$}
		\STATE neg\_subs $\leftarrow$ $\{\neg C| C\in \text{subs}\}$
		\STATE restrictions $\leftarrow$ $\{\}$
		\IF {$|\text{subs}| < k$}
		\STATE fillers $\leftarrow$ $\{\top, \bot, A\}$
		\ELSE
		\STATE $\text{fillers} \leftarrow\{\top, \bot, A\}\cup\textsc{RandSample}(\text{subs}, n=k) \cup \textsc{RandSample}(\text{neg\_subs}, n=k)$
		\ENDIF
		\FOR{$C$ in fillers}
		\FOR{$R$ in role names of $\mathcal{K}$}
		\STATE restrictions $\leftarrow$ restrictions $\cup$ $\{\exists\ R.C\}$
		\STATE restrictions $\leftarrow$ restrictions $\cup$ $\{\forall\ R.C\}$
		\ENDFOR
		\ENDFOR
		\STATE constructs $\leftarrow$ subs $\cup$ neg\_subs $\cup$ restrictions
		\STATE \textbf{return} constructs
	\end{algorithmic}
\end{algorithm}

\begin{algorithm}[p]
	\caption{Function \textsc{RefineAtomicConcept}}
	\label{alg:refine}
	\textbf{Input}: Knowledge base $\mathcal{K}$, atomic concept $A$\\
	\textbf{Hyper-parameters}: Fraction of constructs to be sampled: $\mathit{construct\_frac} \in (0,1]$, default 0.8;\\
	\phantom{\textbf{Hyper-parameters}:} max concept length to be generated: $\mathit{max\_length}$, default 15\\
	\textbf{Output}: Set of concepts $\{C_1, \dots, C_n\}$ which are refinements of $A$
	\begin{algorithmic}[1] 
		\STATE constructs = \textsc{RefineHelper}($\mathcal{K}, A$)
		\STATE $m \leftarrow \lfloor \mathit{construct\_frac} \times \textsc{SizeOf}(\text{constructs}) \rfloor$ \quad \# \textit{Integer part function}
		\STATE constructs $\leftarrow$ \textsc{RandSample}(constructs, $n=m$) \quad \# \textit{Sample without replacement}
		\STATE subs $\leftarrow$ \textsc{Subconcepts}$_\mathcal{K}(A)$ \quad \# \textit{Subconcepts of $A$ in $\mathcal{K}$}
		\STATE result $\leftarrow$ subs \quad \# \textit{All subconcepts are of length 1}
		\FOR{$S_1$ in subs}
		\FOR{$S_2$ in constructs}
		\IF{$S_1\neq S_2$ and $\mathit{length}(S_1 \sqcap S_2) \le \mathit{max\_length}$}
		\STATE result $\leftarrow$ result $\cup$ $\{S_1 \sqcap S_2\}$
		\ENDIF
		\IF{$S_1\neq S_2$ and $S_2$ $\in$ subs and $\mathit{length}(S_1 \sqcup S_2) \le \mathit{max\_length}$}
		\STATE result $\leftarrow$ result $\cup$  $\{S_1 \sqcup S_2\}$ 
		\ELSIF{$S_1\neq S_2$ and $\mathit{length}((S_1 \sqcup S_2)\sqcap A) \le \mathit{max\_length}$}
		\STATE result $\leftarrow$ result $\cup$ $\{(S_1 \sqcup S_2)\sqcap A\}$ \ \# \textit{All refinements are downward refinements}
		\ENDIF
		\ENDFOR
		\ENDFOR
		\STATE \textbf{return} result
	\end{algorithmic}
\end{algorithm}

%% file: eswc22-clp-part6.tex
\section{Evaluation}
\label{sec:evaluation}
\paragraph{Datasets.}
We used four datasets for our experiments: Carcinogenesis, Mutagenesis, Semantic Bible, and Vicodi. All datasets are available in our GitHub repository and described in Table \ref{tab:datasets}. 
We conducted two sets of experiments. First, we wanted to know which neural architecture performs best at predicting concept lengths. Second, we assessed CLIP's performance w.r.t. its runtime and F-measure when compared with the state-of-the-art refinement approaches dubbed CELOE, OCEL, ELTL, and DL-Foil.

\input{table-datasets}

\paragraph{Hardware.}
\label{subsec:hardware}
The training of our concept length learners was carried out on a single 11GB memory NVIDIA K80 GPU with 4 Intel Xeon E5-2670 CPUs at 2.60GHz, and 24GB RAM. During concept learning with CELOE, OCEL, ELTL, and CLIP, we used an 8-core Intel Xeon E5-2695 at 2.30GHz, and 16GB RAM to ensure a fair comparison.

\subsection{Concept Length Prediction}
\label{hyper-parameter}

\subsubsection{Hyper-parameter Optimization.}
In our preliminary experiments on all four knowledge bases, we used a random search \cite{bergstra2012random} to select fitting hyper-parameters (as summarized in Table \ref{tab:hyperparam}). Our experiments suggest that choosing two layers for the recurrent neural networks (LSTM, GRU) is the best choice in terms of computation cost and classification accuracy. In addition, two linear layers, batch normalization, and dropout layers are used to increase the performance. The CNN model consists of two convolution layers, two linear layers, two dropout layers, and a batch normalization layer. Finally, we chose 4 layers for the MLP model with batch normalization and dropout layers. The Rectified Linear Unit (ReLU) is used in the intermediate layers of all models, whereas the sigmoid function is used in the output layers.

We ran the experiments in a $10$-fold cross-validation setting with ten repetitions. Table \ref{tab:hyperparam} gives an overview of the hyper-parameter settings on each of the four knowledge bases considered. The number of epochs was set based on the training speed and the performance of the validation dataset. For example, on the Carcinogenesis knowledge base, most length predictors are able to reach 90\% accuracy with just 50 epochs, which suggests that more epochs would probably lead to overfitting.
\input{table-hyperparameters}
Adam optimizer \cite{kingma2014adam} is used to train the length predictors. We varied the number of examples $\textbf{N}$ between $200$ and $1000$, and the embedding dimension $\textbf{d}$ from $10$ to $100$, but we finally chose $\textbf{N}=\min (1000, \frac{|\mathcal{K}_\mathcal{I}|}{2})$ and $\textbf{d}=40$ as best values for both classification accuracy and computation cost on the four datasets considered.
 
\subsubsection{Results.}
\label{sec:length-prediction}
Table \ref{tab:CLIP-runtime} shows the number of parameters and training time of LSTM, GRU, CNN, and MLP architectures on each of the datasets. From the table, we can observe that our concept length predictors can be trained in less than an hour and be used for efficient concept learning on corresponding knowledge bases.

\begin{figure}
    \centering
    \begin{subfigure}{\textwidth}
    \includegraphics[width=\textwidth]{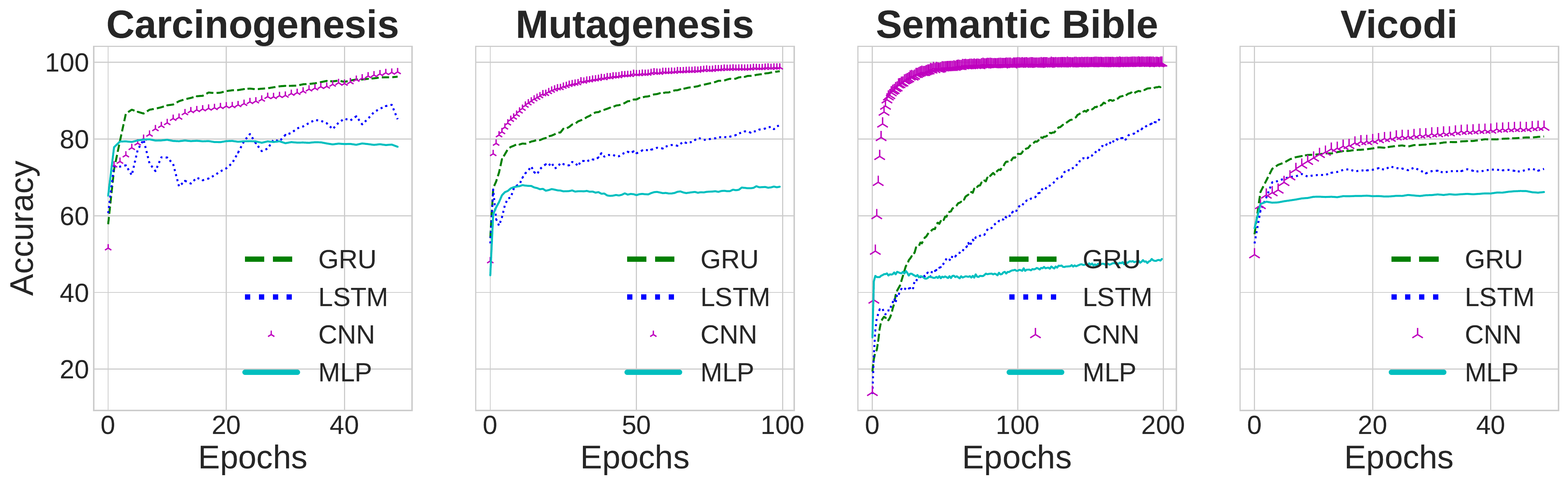}
    \caption{Cross-validation accuracy on the training dataset}
    \label{fig:train-acc}
\end{subfigure}
\vfill
\begin{subfigure}{\textwidth}
    \includegraphics[width=\textwidth]{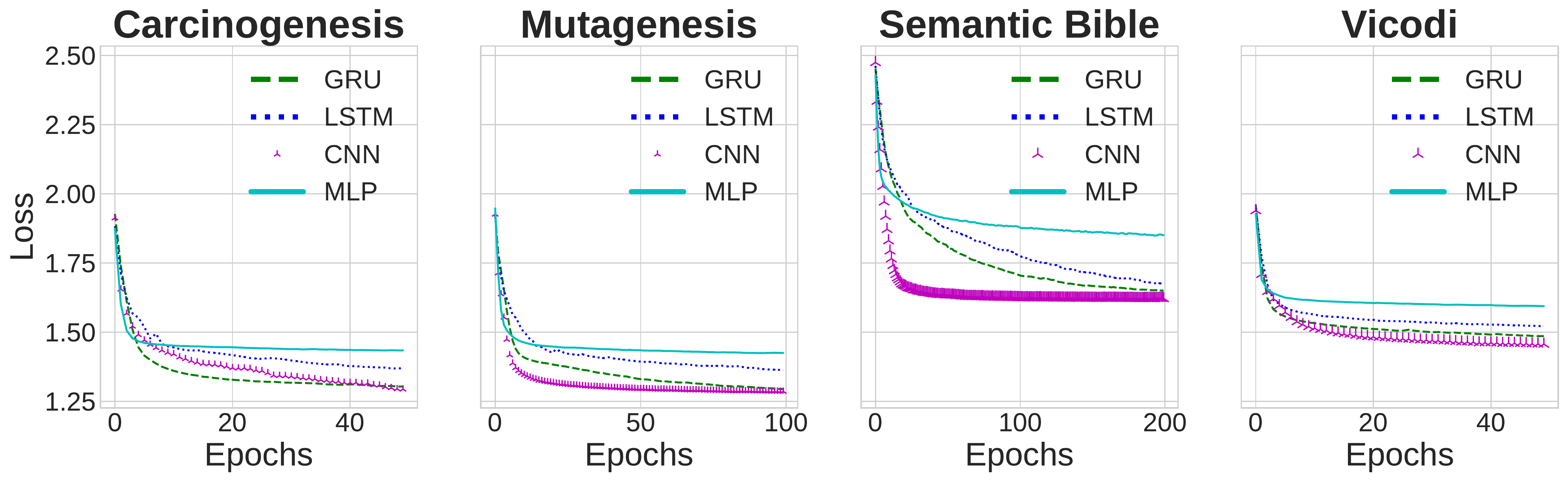}
    \caption{Cross-validation loss on the training dataset}
    \label{fig:train-loss}
\end{subfigure}
\vfill
\begin{subfigure}{\textwidth}
\includegraphics[width=\textwidth]{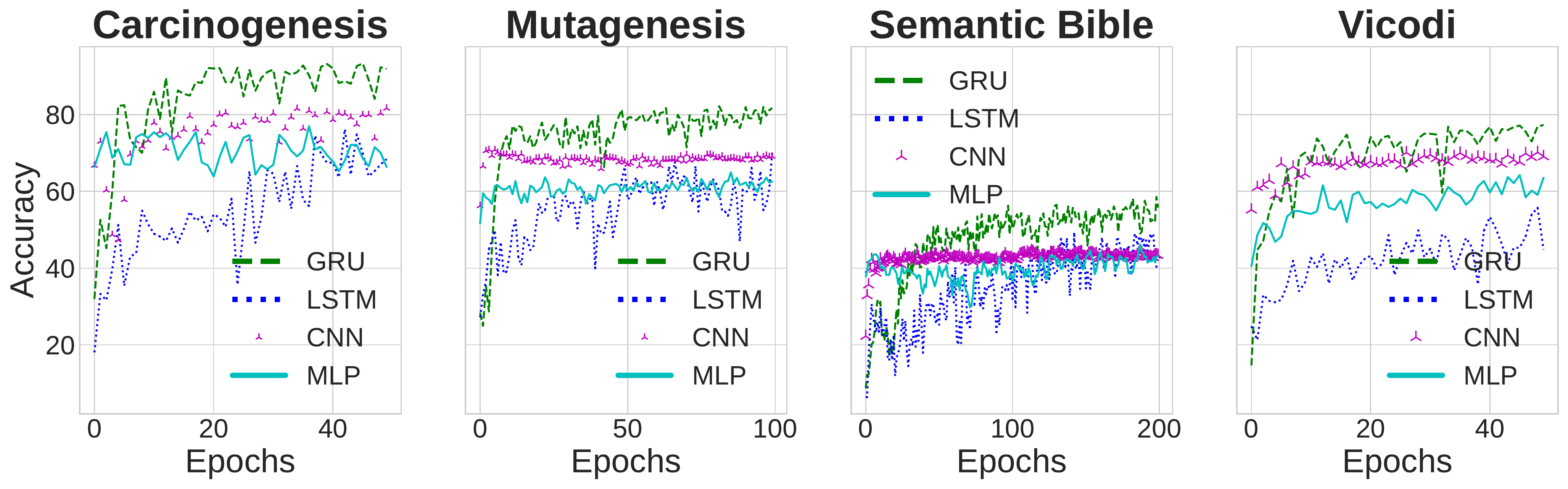}
\caption{Cross-validation accuracy on the validation dataset}
\label{fig:val-acc}
\end{subfigure}
\vfill
\begin{subfigure}{\textwidth}
    \includegraphics[width=\textwidth]{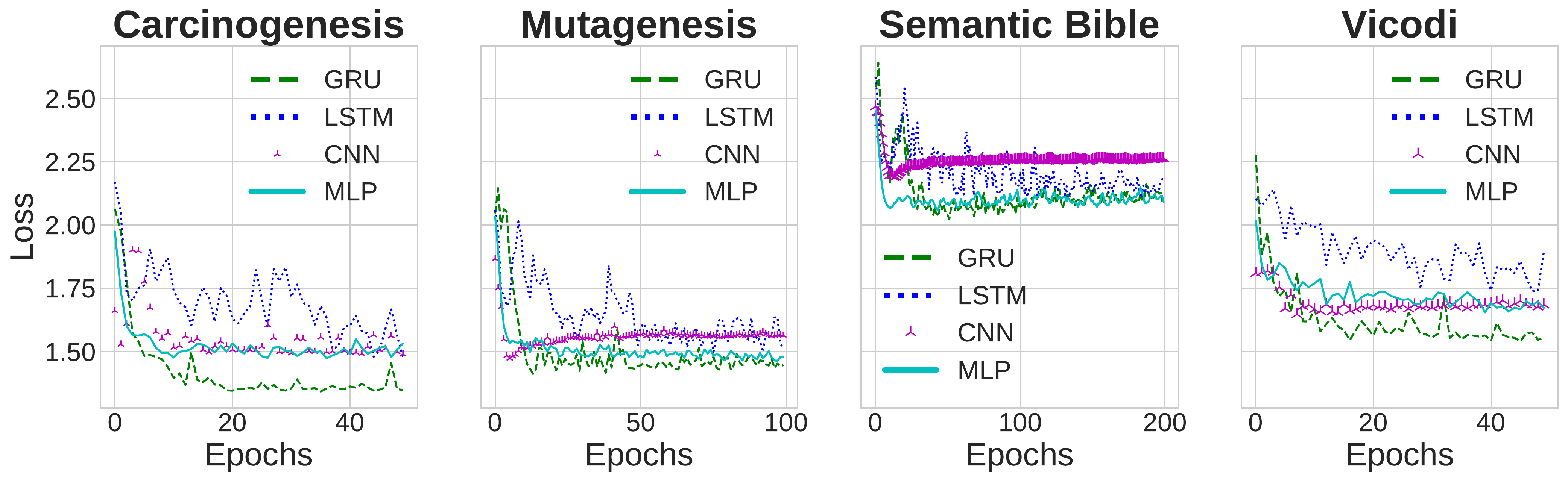}
    \caption{Cross-validation loss on the validation dataset}
    \label{fig:val-loss}
\end{subfigure}
\caption{Training and validation curves}
\label{train-val-curves}
\end{figure}

In Figure \ref{train-val-curves}, we show the training curves for each model on all datasets. 
We can observe a decreasing loss on all knowledge bases (see Figure \ref{fig:train-loss}), which suggests that the models were able to learn. Moreover, the Gated Recurrent Unit (GRU) model outperforms the other models on all datasets, see Figure \ref{fig:val-acc} and Table \ref{tab:length-prediction}. The input to the MLP model is the average of the embeddings of the positive and negative examples for a concept. This may have caused loss of information in the inputs. As shown in Figure~\ref{fig:train-acc}, MLP curves tend to saturate in the early stages of training. We also assessed the element-wise multiplication of the embeddings and obtained similar results. However, as reflected in Table \ref{tab:length-prediction}, all our proposed architectures outperform a random model that knows the distribution of the lengths of concepts in the training dataset. A modified version of MLP where the embedding of each example is processed independently before averaging the final output (no interaction) yielded even poorer results. This suggests that the full interaction between examples is the main factor in the increased performance of recurrent neural networks.

\input{table-evaluation-length-prediction}

Table \ref{tab:length-prediction} compares our chosen neural network architectures and a random model on the Carcinogenesis, Vicodi, Mutagenesis, and Semantic Bible knowledge bases. From the table, it appears that recurrent neural network models (GRU, LSTM) outperform the other two models (CNN and MLP) on three out of four datasets, with the only exception that the LSTM model slightly dropped in performance on Vicodi compared to CNN.
\input{table-evaluation-concept-learning}
While the CNN model tends to overfit on all knowledge bases, the MLP model is unable to extract meaningful information from the average embeddings. On the Semantic Bible knowledge base, which appears to be the smallest dataset, all our proposed networks performed less well than expected. This suggests that our learning approach is more suitable for large knowledge bases. Nonetheless, all our proposed models are clearly better than a distribution-aware random model with a minimum performance (macro F1 score) difference on average between $21.25\%$ (MLP) and $41\%$ (GRU).

\subsection{Concept Learning}
\subsubsection{Experimental Settings.}
The maximal runtime is set to 2 minutes per learning problem.\footnote{The implementations of OCEL and ELTL in the DL-Learner framework, which we used for our experiments, fail to consider the set threshold accurately. Hence, Table \ref{tab:concept-learning} contains values larger than 2 min for these two algorithms.} For all knowledge bases, we generate 100 random learning problems by (1) creating random $\mathcal{ALC}$ concepts $C$ of maximal length $15$, (2) computing the sets of instances $C_\mathcal{P}$ and $C_\mathcal{N}$, (3) providing $C_\mathcal{P}$ and $C_\mathcal{N}$ to each of the approaches, and (4) measuring the accuracy, the F-measure, the runtime, and the length of the best solution generated within the set timeout. We ran all approaches on the same hardware (see Section~\ref{subsec:hardware}). CLIP was configured to use our best concept length predictor (GRU). Note that a predictor is trained for each dataset (see Table~\ref{tab:dataset-concepts}). Also note that we add the ELTL algorithm---a concept learner for the DL $\mathcal{EL}$---to investigate whether our randomly generated concepts are equivalent to concepts in a simpler description logic. 
\subsubsection{Results.}
Table \ref{tab:concept-learning} presents a comparison of the results achieved by CLIP, CELOE, OCEL, and ELTL; results are formatted $\textit{mean}\pm \textit{standard deviation}$. Note that the table does not contain \textsc{DL-Foil} because it could not solve the learning problems that we considered. For instance, the first learning problem on the Semantic Bible knowledge base targets  $\texttt{SonOfGod} \sqcup (\exists\ \texttt{locationOf.StateOrProvince})$. Here, \textsc{DL-Foil} was stuck on the refinement of $\texttt{GeographicLocation}$ with over $5 \times 10^3$ unsuccessful trials. Similar scenarios were observed on other datasets. We also tried running \textsc{DL-Focl}, but it was not possible using the documentation provided.

Our results suggest that CLIP outperforms the other three algorithms in F1 and in runtime on most datasets. The ELTL algorithm appears to be faster than CELOE and OCEL but slower than CLIP. However, its runtime performance stems from the fact that it detects concepts in the DL $\mathcal{EL}$. Since some of our learning problems can only be solved in $\mathcal{ALC}$ or a more expressive DL, ELTL performs poorly in F1 score on all datasets. This result suggests that we do not generate trivial problems.

We used a Wilcoxon Rank Sum test to check whether the difference in performance between CLIP, CELOE, and OCEL was significant. Significant differences are marked with an asterisk. The null hypothesis for our test was as follows: ``the two distributions that we compare are the same". The significance level was $\alpha=0.05$. The performance differences in F1 between CLIP and the other algorithms are significant on 3 out of 4 datasets.
\footnote{Note that we ran OCEL with its default settings and F1 scores are not available.} With respect to runtimes, we significantly outperform all other algorithms on all datasets. Large time differences correspond to scenarios where CLIP detects short solution concepts while other algorithms explore longer concepts. Low time differences correspond to either simple learning problems, where all algorithms find a solution in a short period of time, or complex learning problems where CLIP explores long concepts as other algorithms. 

The average runtimes of CELOE, OCEL, and CLIP across all datasets are $0.85$, $8.08$, and $0.1$ minutes, respectively. Given that the average training time of the length predictor GRU is $4.10$ minutes, we can conjecture the following: (1) the expected number of learning problems from which CLIP should be preferred over CELOE is $5$, and (2) CLIP should be preferred over OCEL for any number of learning problems.

%% file: table-datasets.tex
\begin{table}[tb]
 \centering
  \caption{Overview of benchmark datasets}
  \label{tab:datasets}
  \setlength{\tabcolsep}{2.5pt}
    \begin{tabular}{@{}lcccccc@{}}
        \toprule
        \textbf{Dataset} & $|$\textbf{Individuals}$|$ & $|$\textbf{At. Concepts}$|$ & $|$\textbf{Obj. Prop.}$|$ & $|$\textbf{Data Prop.}$|$ & $|\boldsymbol{\mathit{TBox}}|$ & $|\boldsymbol{\mathit{ABox}}|$ \\
        \midrule
        Carcinogenesis & 22,372 & 142 & \phantom{0}4 & 15 & 138 & \phantom{0}96,757\\
        Mutagenesis & 14,145 & \phantom{0}86 & \phantom{0}5 & \phantom{0}6 &  \phantom{0}82 & \phantom{0}61,965\\
        Semantic Bible  & \phantom{000}724 & \phantom{0}48 & 29 & \phantom{0}9 & \phantom{0}51 & \phantom{00}3,211\\
        Vicodi  & 33,238 & 194 & 10 & \phantom{0}2 & 193 & 149,634\\
        \bottomrule
    \end{tabular}
\end{table}

%% file: table-hyperparameters.tex
\begin{table}[tb]
\centering
\caption{Hyper-parameter setting}
\setlength{\tabcolsep}{6pt}
\begin{tabular}{@{}lccccc@{}}
        \toprule
        \textbf{Dataset} & $|$\textbf{Epochs}$|$ & \textbf{lr} & \textbf{d} & \textbf{Batch Size} & \textbf{N} \\
        \midrule
        Carcinogenesis & \phantom{0}50 & 0.003 & 40 & 512 & 1,000\\
        Mutagenesis & 100 & 0.003 & 40 & 512 &  1,000\\
        Semantic Bible  & 200 & 0.003 & 40 & 256 & \phantom{00}362\\
        Vicodi & \phantom{0}50 & 0.003 & 40 & 512 &  1,000\\
        \bottomrule
    \end{tabular}
\label{tab:hyperparam}
\end{table}
\begin{table}[tb]
\centering
	\caption{Model size and training time}
 	\setlength{\tabcolsep}{6pt}
    \begin{tabular}{@{}lcc@{\hskip 10pt}cc@{}}
		\toprule
	     \textbf{Model} &\multicolumn{2}{c}{\textbf{Carcinogenesis}} &\multicolumn{2}{c}{\textbf{Mutagenesis}}\\
	    \cmidrule(lr){2-3}
		\cmidrule(l){4-5}
		 & $|$\textbf{Parameters}$|$  & \textbf{Train. Time (s)} & $|$\textbf{Parameters}$|$  & \textbf{Train. Time (s)}\\
		\midrule
	    LSTM & 160,208 & 188.42 & 160,208 & 228.13\\
		GRU & 125,708 & 191.16 & 125,708 & 228.68\\
	    CNN & 838,968 & \phantom{0}16.77 & 838,248 & \phantom{0}44.74 \\
	    MLP & \phantom{0}61,681 &  \phantom{0}10.04 & \phantom{0}61,681  & \phantom{0}14.29\\
	    \bottomrule
	    \toprule
	    \textbf{Model} &\multicolumn{2}{c}{\textbf{Semantic Bible}} &\multicolumn{2}{c}{\textbf{Vicodi}}\\
		\cmidrule(lr){2-3}
		\cmidrule(l){4-5}
		 & $|$\textbf{Parameters}$|$  & \textbf{Train. Time (s)} & $|$\textbf{Parameters}$|$  & \textbf{Train. Time (s)}\\
		\midrule
		LSTM & 161,012 & 196.20 & 160,409 & 362.28\\
		GRU & 125,512 & 197.86 & 125,909 & 367.55\\
		CNN & \phantom{0}96,684 & \phantom{0}18.43 & 839,377 & \phantom{0}71.95\\
	    MLP & \phantom{0}61,933 & \phantom{00}9.56 & \phantom{0}61,744  & \phantom{0}24.61\\
		\bottomrule
	\end{tabular}
	\label{tab:CLIP-runtime}
\end{table}

%% file: table-evaluation-length-prediction.tex
\begin{table}[tb]
	\centering
	\caption{Effectiveness of concept length prediction. RM is a random model that makes predictions according to the length distribution in the training dataset, and F1 is the macro F-measure.}
    \label{tab:length-prediction}
	\setlength{\tabcolsep}{4.7pt}
	\footnotesize
    %\resizebox{\textwidth}{!}{
    \begin{tabular}{@{}l@{\hskip 8pt}ccccc@{\hskip 10pt}ccccc@{}}
		\toprule
		\textbf{Metric} &\multicolumn{5}{c}{\textbf{Carcinogenesis}} & \multicolumn{5}{c}{\textbf{Mutagenesis}}\\
		\cmidrule(r){2-6}
		\cmidrule(l){7-11}
		& \textbf{LSTM} & \textbf{GRU} & \textbf{CNN} & \textbf{MLP} & \textbf{RM} & {\hskip 8pt} \textbf{LSTM} & \textbf{GRU} & \textbf{CNN} & \textbf{MLP} & \textbf{RM} \\
		\midrule
		 Train.\ Acc.  & $0.89$ & $0.96$ & $0.97$ & $0.80$ & $0.48$ &{\hskip 8pt} $0.83$ & $0.97$ & $0.98$ & $0.68$ & $0.33$ \\
		Val.\ Acc. & $0.76$ & 
		$0.93$ & $0.82$ & $0.77$ & $0.48$ &{\hskip 8pt} $0.70$ & 
		$0.82$ & $0.71$ & $0.65$ & $0.35$\\
		Test Acc. & $0.92$ & 
		$\mathbf{0.95}$ & $0.84$ & $0.80$ & $0.49$ & {\hskip 8pt}  $0.78$ & 
		$\mathbf{0.85}$ & $0.70$ & $0.68$ & $0.33$\\
		Test F1 & $0.88$ & $\mathbf{0.92}$ & $0.71$  & $0.59$ & $0.33$ &{\hskip 8pt} $0.76$ & $\mathbf{0.85}$ & $0.70$  & $0.67$ & $0.32$\\ 
		\bottomrule
		\toprule
		\textbf{Metric } &\multicolumn{5}{c}{\textbf{Semantic Bible}} &\multicolumn{5}{c}{\textbf{Vicodi}}\\
		\cmidrule(r){2-6}
		\cmidrule(l){7-11}
		& \textbf{LSTM} & \textbf{GRU} & \textbf{CNN} & \textbf{MLP} & \textbf{RM} & {\hskip 8pt} \textbf{LSTM} & \textbf{GRU} & \textbf{CNN} & \textbf{MLP} & \textbf{RM} \\
		\midrule
	    Train. Acc.  & $0.85$ & $0.93$ & $0.99$ & $0.68$ & $0.33$ &{\hskip 8pt} $0.73$ & $0.81$ &  $0.83$ & $0.66$ & $0.28$\\
		Val. Acc. & $0.49$ & 
		$0.58$ & $0.44$ & $0.46$ & $0.26$ &{\hskip 8pt}  $0.55$ & $0.77$ & $0.70$ & $0.64$ & $0.30$\\
		Test Acc. & $0.52$ & 
		$\mathbf{0.53}$ & $0.37$ & $0.40$ & $0.25$ &{\hskip 8pt}  $0.66$ & $\mathbf{0.80}$ & $0.69$ & $0.66$ & $0.29$\\
		Test F1 & $0.27$ & $\mathbf{0.38}$ & $0.20$  & $0.22$ & $0.16$ &{\hskip 8pt} $0.45$ & $\mathbf{0.50}$ & $0.45$ & $0.38$ & $0.20$\\ 
		\bottomrule
	\end{tabular}%}
\end{table}

%% file: table-evaluation-concept-learning.tex
\begin{table}[ht!]
\centering
	\caption{Performance of CLIP compared with CELOE, OCEL, and ELTL on 100 learning problems per knowledge base. The presence of an asterisk indicates that the performance difference is significant between CLIP and the best between CELOE and OCEL. The upward arrow ($\uparrow$) indicates that the higher is better, whereas the downward arrow ($\downarrow$) indicates the opposite. All results are average results per knowledge base. The average time is in minutes. ELTL is shown in gray since it learns concepts in $\mathcal{EL}$ instead of $\mathcal{ALC}$ as the others do.}
	\label{tab:concept-learning}
    \scriptsize
    \begin{tabular}{l c c c c c
					>{\color{gray}}cc
					>{\columncolor[gray]{0.9}}c}
				\toprule
				\textbf{Metric} & \multicolumn{7}{c}{\textbf{Carcinogenesis}}\\
				\cmidrule{2-9}
			    && \textbf{CELOE} && \textbf{OCEL}  && \textbf{ELTL} && \textbf{CLIP}\\
				\midrule
				Acc. $\uparrow$ && $\phantom{*}0.78 \pm 0.27$ && $\phantom{*}0.89 \pm 0.31$ && $\phantom{*}0.58 \pm 0.46$ && $ \mathbf{0.99} \phantom{*} \pm 0.00$\\
				
				F1 $\uparrow$ && $\phantom{*}0.62 \pm 0.46$ && $\phantom{*}-$ && $\phantom{*}0.51 \pm 0.47$ && $\mathbf{0.96*} \pm 0.10$ \\
				
				Runtime (min) $\downarrow$ && $\phantom{*}0.93 \pm 0.94$ && $\phantom{*}3.01 \pm 0.72$ && $\phantom{*}0.75 \pm 0.07$ && $ \mathbf{0.10*} \pm 0.09$ \\
				
				Length $\downarrow$ && $\phantom{*}\mathbf{1.69} \pm 0.89$ && $\phantom{*}7.81 \pm 6.88$ && $\phantom{*}1.04 \pm 0.39$ && $\ 2.00 \phantom{*}\pm 1.28$ \\
				\bottomrule
				\toprule
				
				\textbf{Metric} & \multicolumn{7}{c}{\textbf{Mutagenesis}}\\
				\cmidrule{2-9}
				Metric	&& \textbf{CELOE} && \textbf{OCEL}  && \textbf{ELTL} && \textbf{CLIP}\\
				\midrule
				Acc. $\uparrow$ && $\phantom{*}0.99 \pm 0.00$ && $\phantom{*}0.71 \pm 0.45$ && $\phantom{*}0.37 \pm 0.43$ && $ \mathbf{0.99} \phantom{*}\pm 0.00$\\
				
				F1 $\uparrow$ && $\phantom{*}0.81 \pm 0.35$ && $\phantom{*}-$ && $\phantom{*}0.29 \pm 0.40$ && $\mathbf{0.93*} \pm 0.18$ \\
				
				Runtime (min) $\downarrow$ && $\phantom{*}0.70 \pm 0.77$ && $\phantom{*}2.39 \pm 0.18$ && $\phantom{*}0.29 \pm 0.16$  && $ \mathbf{0.07*} \pm 0.05$ \\
				
				Length $\downarrow$ && $\phantom{*}2.79 \pm 1.17$ && $12.63 \pm 7.03$ && $\phantom{*}1.10 \pm 0.81$ && $ \mathbf{2.20} \phantom{*} \pm 1.16$ \\
				\bottomrule
				\toprule
				
				\textbf{Metric} & \multicolumn{7}{c}{\textbf{Semantic Bible}}\\
				\cmidrule{2-9}
				&& \textbf{CELOE} && \textbf{OCEL}  && \textbf{ELTL} && \textbf{CLIP}\\
				\midrule
				Acc. $\uparrow$ && $\phantom{*}0.99 \pm 0.02$ && $0.66 \pm 0.47$ && $\phantom{*}0.59 \pm 0.37$ && $ \mathbf{0.99} \phantom{*} \pm 0.00$\\
				
				F1 $\uparrow$ && $\phantom{*}0.97 \pm 0.10$ && $-$ && $\phantom{*}0.57 \pm 0.38$ && $ \mathbf{0.98} \phantom{*} \pm 0.05$ \\
				
				Runtime (min) $\downarrow$ && $\phantom{*}0.47 \pm 0.80$ && $22.15 \pm 96.55$ && $\phantom{*}0.09 \pm 0.07$ && $ \mathbf{0.06*} \pm 0.05$ \\
				
				Length $\downarrow$ && $\phantom{*}3.85 \pm 2.44$ && $9.54 \pm 5.73$ && $\phantom{*}1.38 \pm 1.76$ && $ \mathbf{2.52*} \pm 1.26$ \\
				\bottomrule
				\toprule
				
				\textbf{Metric} & \multicolumn{7}{c}{\textbf{Vicodi}}\\
				\cmidrule{2-9}
				&& \textbf{CELOE} && \textbf{OCEL}  && \textbf{ELTL} && \textbf{CLIP}\\
				\midrule
				Acc. $\uparrow$ && $\phantom{*}0.29 \pm 0.44$ && $\phantom{*}0.25 \pm 0.43$ && $\phantom{*}0.28 \pm 0.44$ && $\mathbf{0.99*} \pm 0.00$\\
				
				F1 $\uparrow$ && $\phantom{*}0.25 \pm 0.44$ && $\phantom{*}-$ && $\phantom{*}0.25 \pm 0.44$ && $\mathbf{0.97*} \pm 0.09$ \\
				
				Runtime (min) $\downarrow$ && $\phantom{*}1.30 \pm 0.71$ && $\phantom{*}4.78 \pm 1.12$ && $\phantom{*}1.81 \pm 0.46$ && $\mathbf{0.16*} \pm 0.12$ \\
				
				Length $\downarrow$ && $10.79 \pm 6.30$ && $11.54 \pm 6.00$ && $11.14 \pm 6.11$ && $\mathbf{1.68*} \pm 0.98$ \\
				\bottomrule
		\end{tabular}
\end{table}

%% file: eswc22-clp-sum.tex
\section {Conclusion and Future Work}
\label{sec:conclusion}
We investigated the prediction of concept lengths in the description logic $\mathcal{ALC}$, to speed up the concept learning process using refinement operators. To this end, four neural network architectures were evaluated on four benchmark knowledge bases. The evaluation results suggest that all of our proposed models are superior to a random model, with recurrent neural networks performing best at this task. We showed that integrating our concept length predictors into a concept learner can reduce the search space and improve the runtime and the quality (F-measure) of solution concepts.

Even though our proposed learning approach was very efficient when dealing with concepts of length up to 11 (in $\mathcal{ALC}$), its behavior is not guaranteed when longer concepts are considered. Moreover, the use of generic embedding techniques might lead to suboptimal results. In future work, we plan to jointly learn the embeddings of a given knowledge graph and the lengths of its complex (long) concepts. We will also explore further network architectures such as multi-set convolutional networks~\cite{zaheer2017deepsets} and neural class expression synthesis~\cite{kouagou2021neural}.

\subsubsection*{Acknowledgements.}
This work is part of a project that has received funding from the European Union's Horizon 2020 research and innovation programme under the Marie Skłodowska-Curie grant agreement No 860801. This work has been supported by the German Federal Ministry of Education and Research (BMBF) within the project DAIKIRI under the grant no 01IS19085B and by the German Federal Ministry for Economic Affairs and Climate Action (BMWK) within the project RAKI under the grant no 01MD19012B. The authors gratefully acknowledge the funding of this project by computing time provided by the Paderborn Center for Parallel Computing (PC\textsuperscript{2}).